\documentclass[conference,10pt]{IEEEtran}
\usepackage{epsfig,rotating,setspace,latexsym,amsmath,epsf,amssymb,amsfonts,bm,theorem,subfigure,epstopdf}
\usepackage{cite,authblk}
\usepackage{bbm}
\usepackage{color}
\usepackage{mathtools}
\usepackage{algorithm}
\usepackage{algpseudocode}
\usepackage{verbatim}
\usepackage{xcolor}

\algrenewcommand\algorithmicforall{\textbf{foreach}}
\algrenewcommand\algorithmicindent{.8em}

\algnewcommand\algorithmicforeach{\textbf{for each}}
\algdef{S}[FOR]{ForEach}[1]{\algorithmicforeach\ #1\ \algorithmicdo}

\IEEEoverridecommandlockouts
\allowdisplaybreaks

\begin{document}

\title{Semantic Multi-Resolution Communications}

\author{
Matin Mortaheb$^{\dag}$, Mohammad A. (Amir) Khojastepour$^{*}$, Srimat T. Chakradhar$^{*}$, Sennur Ulukus$^{\dag}$ \\
\normalsize $^{\dag}$University of Maryland, College Park, MD, $^{*}$NEC Laboratories America, Princeton, NJ \\
\normalsize \emph{mortaheb@umd.edu, amir@nec-labs.com, chak@nec-labs.com, ulukus@umd.edu}
}

\maketitle

\begin{abstract}
Deep learning based joint source-channel coding (JSCC) has demonstrated significant advancements in data reconstruction compared to separate source-channel coding (SSCC). This superiority arises from the suboptimality of SSCC when dealing with finite block-length data. Moreover, SSCC falls short in reconstructing data in a multi-user and/or multi-resolution fashion, as it only tries to satisfy the worst channel and/or the highest quality data. To overcome these limitations, we propose a novel deep learning multi-resolution JSCC framework inspired by the concept of multi-task learning (MTL). This proposed framework excels at encoding data for different resolutions through hierarchical layers and effectively decodes it by leveraging both current and past layers of encoded data. Moreover, this framework holds great potential for semantic communication, where the objective extends beyond data reconstruction to preserving specific semantic attributes throughout the communication process. These semantic features could be crucial elements such as class labels, essential for classification tasks, or other key attributes that require preservation. Within this framework, each level of encoded data can be carefully designed to retain specific data semantics. As a result, the precision of a semantic classifier can be progressively enhanced across successive layers, emphasizing the preservation of targeted semantics throughout the encoding and decoding stages. We conduct experiments on MNIST and CIFAR10 dataset. The experiment with both datasets illustrates that our proposed method is capable of surpassing the SSCC method in reconstructing data with different resolutions, enabling the extraction of semantic features with heightened confidence in successive layers. This capability is particularly advantageous for prioritizing and preserving more crucial semantic features within the datasets.
\end{abstract}
 
\section{Introduction}
In traditional communication systems, separate source and channel coding (SSCC) is utilized to express the data as efficiently as possible and to mitigate the effects of noise and interference on the communication channel, respectively. An efficient source encoder generates minimal encoded representation of the source data for a given distortion metric. On the other hand, reliable channel encoding can be achieved by expanding the data via adding redundancies, ensuring that the transmitted data can be decoded by the receiver with negligible error.
While Shannon's source-channel separation theorem \cite{shannon1948mathematical} exists for a single link, SSCC is not optimal for finite block lengths and/or for multi-user scenarios \cite{gunduz2009source}. Indeed, many recent results show the improvement of JSCC over SSCC in such scenarios \cite{kurka2020deepjscc, yang2021deep, yang2023witt}.

Sub-optimality of SSCC in multi-user scenarios, e.g., multi-casting, is further imposed by a limited channel coding rate to support the user with the worst channel. Hence, the users with the better channel will encounter the same reconstruction performance (RP) as the user with the weakest channel. In JSCC, sacrificing a small portion of RP for weaker users can result in larger gains in RP for better users. This trade-off allows for better overall reconstruction performance across all users. Recently deep learning techniques such as auto-encoder (AE) \cite{baldi2012autoencoders} is introduced to perform JSCC efficiently in an end-to-end communication system \cite{aoudia2019model, aoudia2021end}. Considering the finite length coding, the drop in weak user RP in comparison to SSCC is shown to be negligible to none in practice.

The idea of multi-resolution (MR) communications is motivated by scenarios such as multimedia-communication where different receivers desire or require different RPs (\emph{defined as tiers}) or when a receiver wishes to improve its RP successively. For example, consider a multi-casting scenario where users are registered to different tiers of RPs. Without MR communications, (a) the transmitter can individually encode the media for each tier and send it using the supported channel rate of the weakest user in that tier, or (b) encode the media with the highest resolution at the best tier and send it with the supported rate of the overall weakest user, in order to satisfy all users' RP requirements and channel rates. 
In contrast, the MR encoder is designed to provide encoded data for the first (worst) tier and successive (or differential) encoded data for each improved tier. In other words, instead of re-encoding the media for each tier separately, the MR encoder provides differential (or successive) data with considerably smaller sizes to improve the RP of the next better tier. Hence, by using MR encoding, (c) the transmitter is only required to send the successive data for each improved tier at the supported rate of the weakest user in that tier.

A simple calculation may better put the use case of the concept of MR communication in multi-casting in perspective. Consider 4 tiers of RPs where the encoded block size of the tiers are $B, 2B, 4B,$ and $6B$ symbols and the supported rate (for error-free communication) of the lowest user in each tier is $1/3, 1/2, 2/3,$ and $ 3/4$, respectively. Assuming that the MR encoder for each successive layer generates differential output of size $\eta B$, $\eta \geq 1$, the transmission scheme (a), (b), and (c) in the above require transmission of $21B, 18B,$ and $(3+7.66\eta)B$ symbols, respectively. In practice $\eta$ is very close to one, say, $\eta = 1.05$ for which $(3+7.66\eta)B = 11.04B$.

The works \cite{liu2023multiresolution, kurka2019successive} investigate end-to-end multi-resolution communication. In \cite{liu2023multiresolution}, the authors introduce the multi-resolution convolutional AE (MrCAE) approach, which initially processes down-sampled data to capture large-scale, low-dimensional structures. Subsequently, both the data and the neural network undergo iterative refinement using transfer learning at each stage. However, a challenge still exists in transmitting high-resolution data, as the encoder needs to send the encoded data from the beginning. In \cite{kurka2019successive}, the authors propose a solution by dividing the encoded data into layers, and each layer's decoder is designed to utilize the encoded data from previous layers. This technique allows for efficient transmission of encoded data until a certain level of reconstruction quality is achieved, addressing the communication concern more effectively. However, simply dividing the encoded data may not be the best approach in generating multi-resolution representation of the input data.

In this paper, we introduce semantic multi-resolution communication (SMRC), where the goal is not only to provide successive refinement of the RP but also to preserve hierarchical semantic information with finer accuracy through successive encoded outputs. In semantic communication, the goal is to preserve an intended meaning (or semantic) in the encoded data which is otherwise compressed to satisfy a certain RP or distortion value \cite{sagduyu2022semantic, sagduyu2023vulnerabilities}. In SMRC, the encoded output block from the encoder is composed of sub-blocks where each subsequent sub-block provides further information to the decoder in order to (i) improve RP, or (ii) improve the accuracy of a semantic feature or group of semantic features in terms of recall/precision, or both (i) and (ii). In other words, the decoder input consists of a sequential subset of sub-blocks starting from the first sub-block in order to achieve a particular RP and semantic accuracy. Exemplary scenarios discussed in Section~\ref{sec:MH_AE_SMRC} provide more details and better understanding of the possible use-cases of SMRC. 

The main contributions of our paper are as follows: (i) We introduce a semantic multi-resolution communication (SMRC) framework, which enables end-to-end communication with various resolutions while preserving semantic features with different accuracy levels (see Section~\ref{sec:MH_AE_SMRC}). (ii) Inspired by vast literature on multi-task learning, we propose a multi-head AE structure which proves to be effective in realizing SMRC (see Section~\ref{sec:prob_formulation}). (iii) We show the improvement and efficiency of the SMRC framework as well as the effectiveness of the proposed multi-head AE structure through numerical evaluations. We conduct experiments using the MNIST and CIFAR10 datasets to train the models and evaluate the performance of the proposed scheme (see Section~\ref{sec:Exp_section}). 

\section{Multi-Head AE for SMRC} \label{sec:MH_AE_SMRC}

In the following, we first consider two exemplary scenarios for SMRC. Then, we introduce the general SMRC structure which is built based on the concept of multi-task learning.

\begin{figure}[t]
 \centerline{\includegraphics[width=1\linewidth]{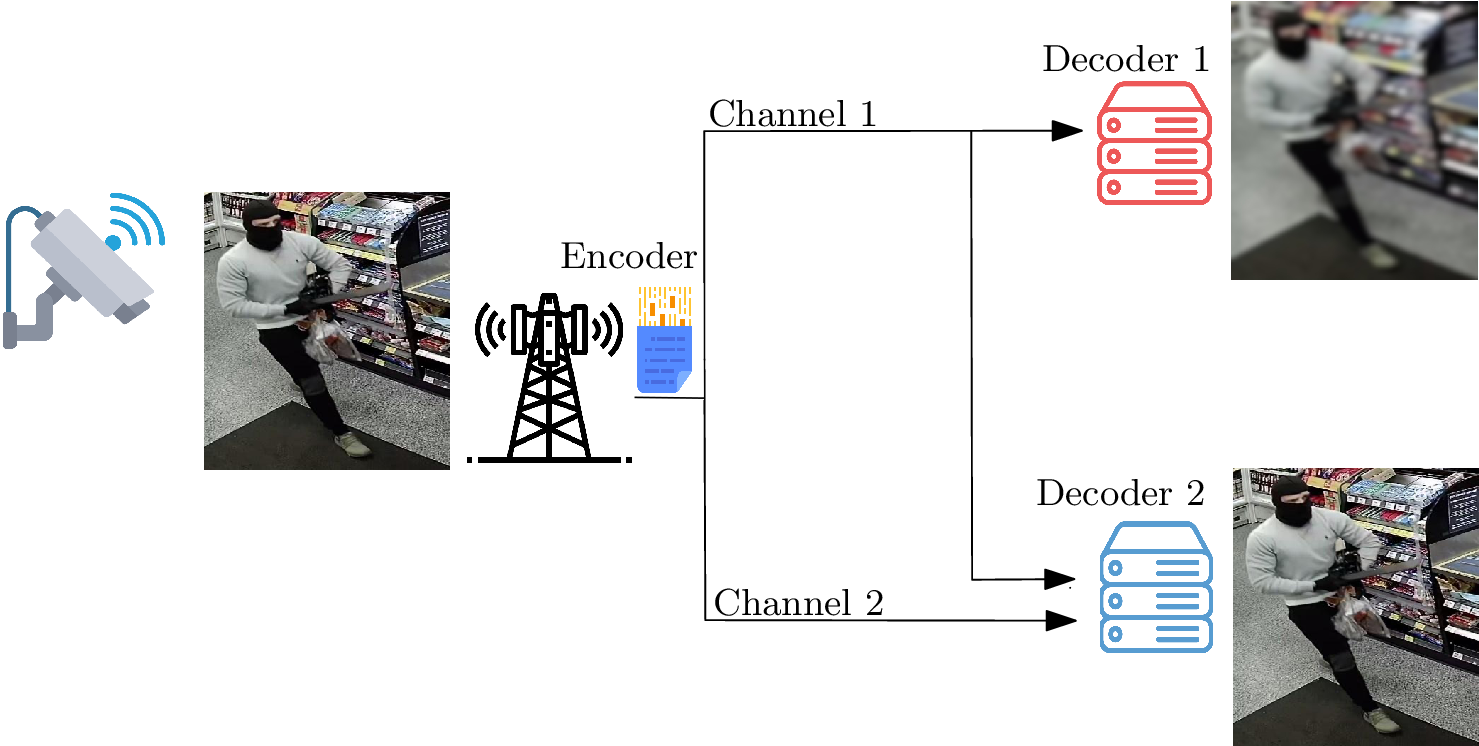}}
  \caption{Application of SMRC on Example 1.}
  \label{EX1}
  \vspace*{-0.4cm}
\end{figure}

\textbf{Example 1: \label{example_1}} Consider a scenario where the goal is bandwidth-efficient transmission of the CCTV video of a shop to the server, as depicted in Fig.~\ref{EX1}. The received signal is not only used by the server to reconstruct the video but also to accurately detect instances of particular emergency situations such as burglary, break-ins, or fire. By employing semantic communication, one can effectively minimize the occurrence of semantic errors along with the reconstruction loss, thereby ensuring the successful preservation of the meaning conveyed by the retrieved information. \cite{sagduyu2022semantic, sagduyu2023vulnerabilities}

In normal conditions, a lower-resolution video is sufficient as long as the semantic information, e.g., a possible emergency event, is preserved. However, when the receiver detects an emergency situation such as burglary, one may need to (i) acquire the higher resolution of the event in order to identify specific details, e.g., a human face or (ii) confirm certain semantic information with higher confidence, e.g., with certain recall (probability of detection) and precision (false alarm probability). 
We note that increasing recall is beneficial to minimize the loss associated with the theft, while precision is important when an incorrect detection incur considerable costs, e.g., when calling 9-1-1 police patrol or other emergency units such as firefighters.  
In such cases, instead of re-sending the entire data from the beginning, our proposed SMRC framework allows for the transmission of successive layers, specifically designed to facilitate the efficient exchange of content with varying resolutions and semantic information with a desired confidence level.

\begin{figure}[t]
 \centerline{\includegraphics[width=1\linewidth]{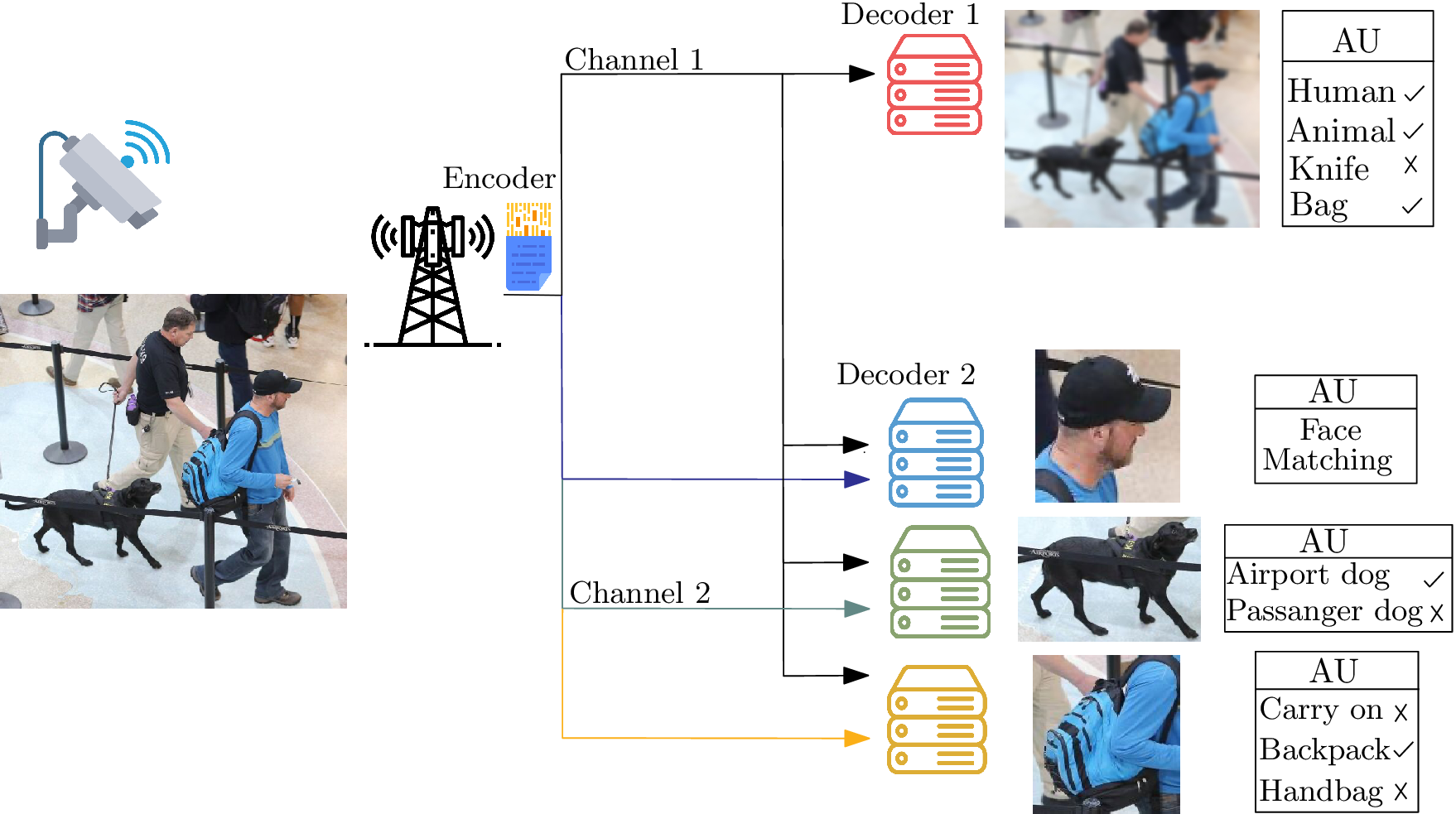}}
  \caption{Application of SMRC on Example 2.}
  \label{EX2}
  \vspace*{-0.4cm}
\end{figure}

\textbf{Example 2: \label{example_2}} Fig.~\ref{EX2} illustrates an alternative scenario where a 5G wireless CCTV camera at the airport terminal gate captures videos of the cars, humans, animals, luggage, etc. The video clips are encoded with a lower resolutions to preserve the bandwidth, nonetheless, encoding is performed to preserve certain semantic information required for running analytic units such as human-detection, animal-detection, and object-detection. Based on a situation, either from time-to-time and at random, or to detect suspicious activities the server-side analytic units may ask for higher resolution video clips, and/or better accuracy of some semantic features, e.g., to run and obtain an improved analytical result, e.g., human-segmentation, object-segmentation, or human-face-recognition. Moreover, for some semantic features, it might be desirable to receive higher resolution video clips, e.g., human-face-detection. The concept of SMRC is helpful in providing incremental information for improving the semantic accuracy and resolution of the reconstruction.

SMRC is composed of an encoder and a decoder. The encoder takes an input block of symbols and produces an encoded block (output block) that is composed of, say $L$, sub-blocks indexed sequentially from 1 to $L$. The decoder takes the sequence of sub-blocks 1 to $l \in [L]$ to generate layer $l$ output that achieves a particular resolution determined by the reconstruction performance (RP) and semantic accuracy (SA). The $(l+1)$th sub-block provides differential information to successively improve the resolution of the layer $l$ output in terms of RP and SA. Naturally, the resolution defined for different output levels should follow a nested structure. This means that a well-defined layer $(l+1)$ output has better or equal resolution than the layer $l$ output. Nonetheless, the desired RP for an input block might be related to the semantic content of the block. This means that the RP for layer $(l+1)$ can be defined in terms of the semantic content obtained in layer $l$ output.  For example, suppose that the semantic preserved in layer $l$ allows to run object-detection, animal-detection, and human-detection analytics and find the corresponding bounding boxes or region-of-interests (RoIs). Then, the RP for the layer $(l+1)$ may be defined to assign different RP for each of RoIs based on their types, i.e., object, human, and animal. 

We also note that the nested structure allows to define a single accuracy metric at the layer $l$ for a group of classes that belong to a semantic type and then further refine the accuracy in the subsequent output levels. This means that one can define a hierarchical structure for semantic types and classes such that all possible cases for the layer $(l+1)$ are subsets of the cases defined for layer $l$. The concept of grouping some classes into a single combined class is called \emph{class pooling}. Hence, the hierarchical structure may be defined by allowing class pooling in layer $l$ based on the classes in layer $(l+1)$, i.e., having a combined class in layer-$l$ by grouping some classes from  layer $(l+1)$. Besides the application of class pooling in defining hierarchical semantic structure, we will later reveal a powerful property of class pooling in joint optimization of recall and precision. 

\begin{figure}[t]
 \centerline{\includegraphics[width=0.9\linewidth]{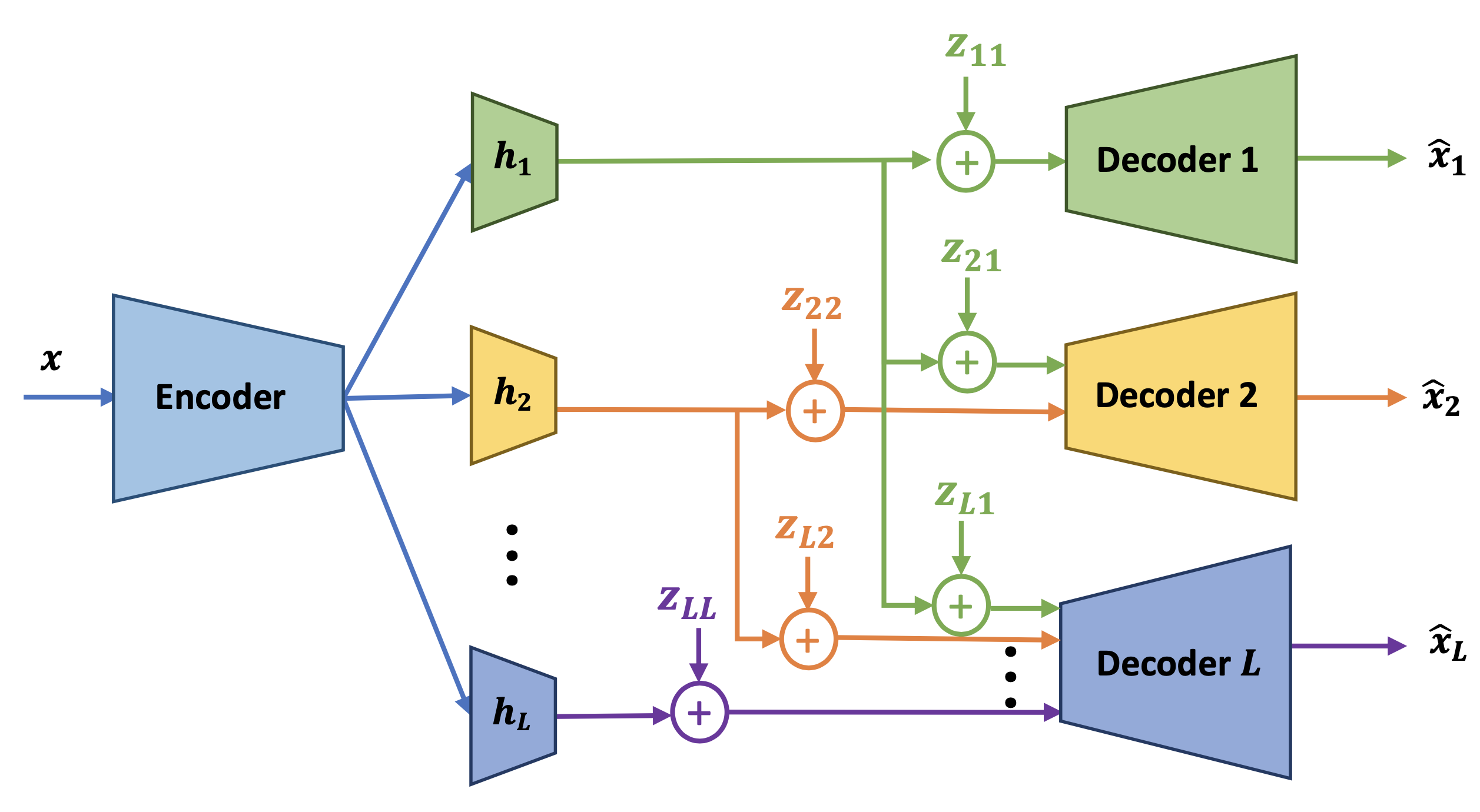}}
  \caption{Multi-head AE for SMRC.}
  \label{system_model}
  \vspace*{-0.4cm}
\end{figure}

Inspired by the vast literature on multi-task learning (MTL) \cite{Zhang2017ASO}, we adopt a multi-head encoder for learning the related tasks of finding the encoding for each sub-block. It is known that MTL is a powerful approach in learning multiple related tasks together and discovering a shared representation in the process. The fundamental idea behind MTL is that data used for these tasks can have underlying similarities that can be captured in a common representation, and not only the shared representation but also the individual task performance is improved when they are learned jointly. The MTL is composed of a network's pre-layer that is shared among all tasks, while task-specific layers are added on top of the pre-layers to perform the function that is specific to each task. To train the model, the overall loss is computed as a weighted sum of individual task losses, where the weights correspond to the importance of each task. This framework allows for efficient and effective learning of multiple tasks simultaneously, leveraging shared information while accommodating task-specific differences. 

Fig.~\ref{system_model} illustrates the general multi-head structure of our proposed SMRC. By enhancing the existing end-to-end multi-resolution framework introduced by \cite{kurka2019successive}, we incorporate a multi-head structure after the encoder stage. The primary objective of SMRC is to efficiently encode data in a hierarchical manner, enabling the decoders to progressively enhance data reconstruction in subsequent levels by leveraging the encoded data from previous layers. Additionally, the multi-head structure in SMRC allows the transmitter to encode the data in order to preserve semantic features independently for each layer, thereby facilitating effective semantic communication. We note that the head $l \in [L]$ produces the sub-block $l$ in the encoded block.  

We normalize the encoder output such that the average power of the transmitted signal is 1. Hence, the SNR of the links is directly controlled by the variance of the noises $z_{li}$ for each link between an encoder head $l$ and decoder $i$ as illustrated in Fig.~\ref{system_model}. This SNR can also be interpreted as the overall signal-to-noise ratio due to the combined effect of channel gain and additive Gaussian noise at the receiver. This can be used to adapt our SMRC model in a number of different multi-user scenarios and channel fading conditions. 

\textbf{Single AP to multiple users, block fading:} Based on this interpretation, Fig.~\ref{system_model} illustrates a multi-user scenario in which a single access point (AP) communicates with multiple users where the encoded output for each head is sent as one packet of transmission and the channel between the AP and each user is block fading, i.e., the channel remains constant during transmission of each packet and may change between transmission of packets. 

\textbf{Single AP to multiple users, AWGN channel:} If the coherence time of the channel covers the entire transmission of the output from all encoder heads, the channel can be modeled as AWGN channel within such transmission and we have $z_{li} = z_{lj}, \forall i,j \in [L]$. In the multi-user interpretation of the model in Fig.~\ref{system_model}, the users are served with different resolutions by transmitting varying numbers of packets corresponding to different heads.

\textbf{Single AP to a single user, block fading:} On the other hand, The model Fig.~\ref{system_model} can be used for multi-resolution transmission of data to a single user by setting $z_{ki} = z_{li}, \forall k,l \in [L]$. In this scenario, the encoded data packet generated by each head is successively transmitted to the user which reproduces the data after reception of the first packet and successively improves its reproduction quality by receiving the next packet. 

\section{System Model and Problem Formulation} \label{sec:prob_formulation}

\begin{figure}[]
 \centerline{\includegraphics[width=0.9\linewidth]{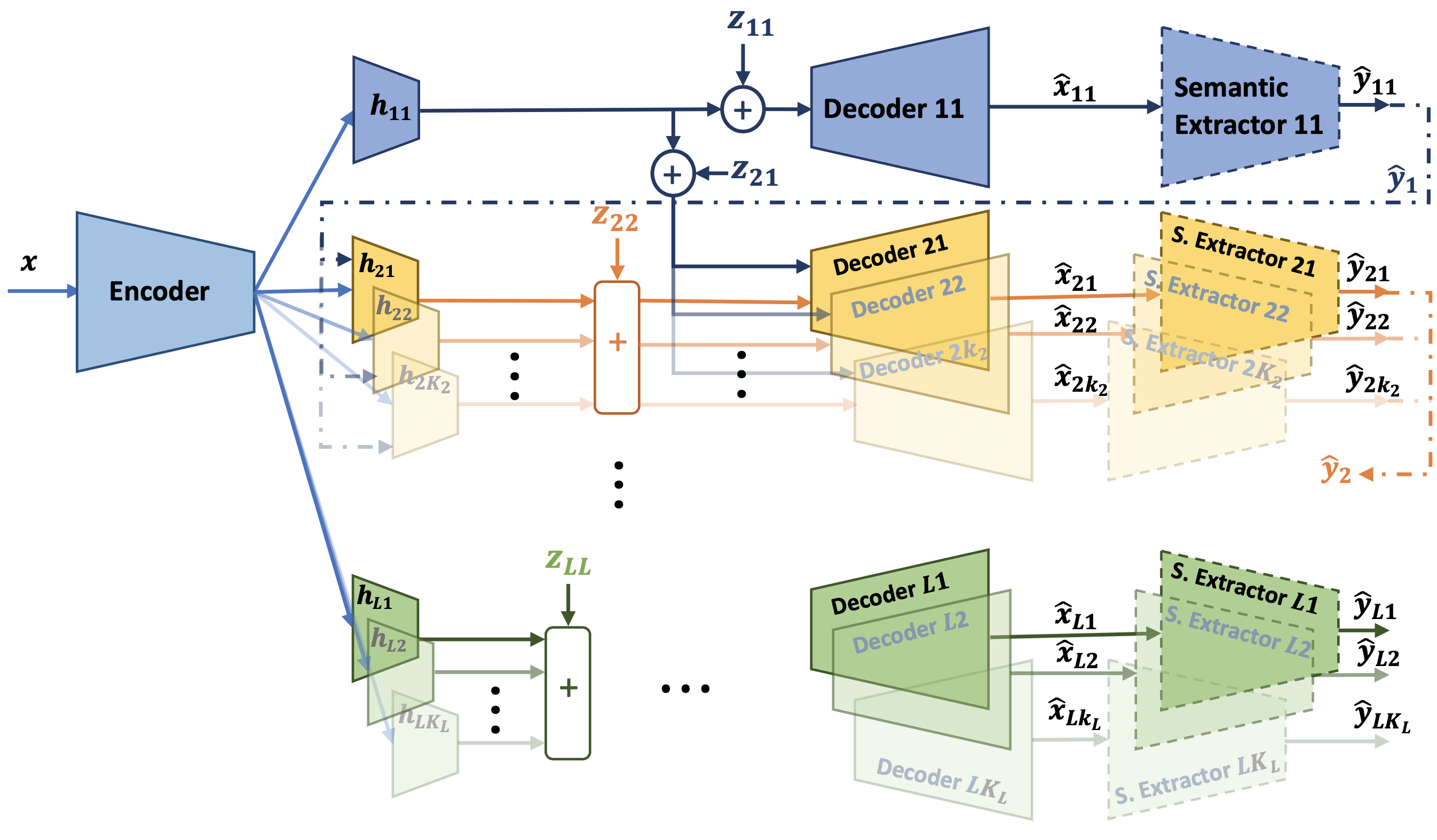}}
  \caption{System model.}
  \label{system_model_2}
  \vspace*{-0.4cm}
\end{figure}

We consider an end-to-end semantic communication system that incorporates a deep learning model as depicted in Fig.~\ref{system_model_2}, consisting of an encoder that produces a semantic multi-resolution encoding of an input data block, denoted as $x$. The output block generated by the encoder consists of $L$ sub-blocks, indexed as $l = 1, 2, \ldots, L$, each containing $n_l$ symbols, where $1 \leq l\leq L$.
The encoder exploits a neural network with a multi-head structure.
The sub-block $l$ is the output of layer $l$ which is generated by one of the encoder heads, say $h_{lk}$, that belongs to layer $l$. This means that there are $K_l$ possible outputs by layer $l$ corresponding to one of its head $h_{lk}, 1 \leq k \leq K_l$ where $K_l$ is the total number of heads in layer $l$.

In reference to Fig.~\ref{system_model_2}, the encoder is responsible for performing joint source and channel encoding of the input data $x$ and the decoder performs both channel and source decoding. The decoder first reconstructs $\hat{x}$ from the received packets transmitted through the communication channel, and then the semantic extractor takes out the semantics $\hat{y}$ from this reconstructed data.
The loss function for each link between an encoder head $h_{lk}$ to the receiver is formulated as
\begin{align} \label{eq:loss}
    \mathcal{L}_{lk} & = \alpha_{lk} \overbrace{||\mathcal{R}_{\hat{Y}_{l-1}} (x-\hat{x}_{lk})||^2}^\text{MSE loss} + \nonumber \\
    & \quad (1-\alpha_{lk}) \underbrace{(\mathbbm{1}^T(y_{lk}(x)) \bm{\beta}_{lk} \log(\bm{\sigma}(\hat{y}_{lk})))}_\text{semantic loss}  
\end{align}
where $\mathbbm{1}(y_{lk}(x))$ is a one-hot column vector obtained from the class label $y_{lk}(x)$, $\bm{\beta}_{lk}$ is a diagonal matrix that models varying degrees of accuracy for different classes, and $\bm{\sigma}$ is a softmax operator. $\hat{Y}_{l-1} = [\hat{y}_1, \ldots, \hat{y}_{l-1}]$ where $\hat{y}_l = [\hat{y}_{l1}, \ldots, \hat{y}_{l,K_l}], \forall l \in [L-1]$. 
$\mathcal{R}_{\hat{Y}_{l-1}}$ is the region of interest (RoI) operator defined as
$\mathcal{R}_{\hat{Y}_{l-1}}(x) = M \odot x$ where $M$ is the mask with entries $m$, $0 \leq m \leq 1$ that is element-wise multiplied by $x$. 
The entries of the mask are used to enforce the reconstruction importance of the corresponding element of $x$. The subscript of the RoI operator $\mathcal{R}_{\hat{Y}_{l-1}}$ denotes the possible dependency of RoI operator to the semantics extracted by previous layers, namely $Y_{l-1}$. For example, the RoIs in Example~2, correspond to a segment or bounding box which includes a human face, an animal face, etc. which may depend on the semantic extractor output of the previous layer.

The loss function \eqref{eq:loss} is defined to jointly minimize the reconstruction loss and semantic loss. The parameter $\alpha_{lk} \in [0,1]$ is used to prioritize reconstruction versus semantic loss. In extreme cases, to maximize the priority of reconstruction loss (semantic loss), we set $\alpha_{lk} = 1$ ($\alpha_{lk} = 0$). For example, when through the reception of the first layer one has identified that a burglary has happened in Example~1, the priority for the second layer might be to increase the resolution of the image in order to identify the face of the burglar in which case we set $\alpha_{21} = 1$ in designing the encoder head $h_{21}$. Conversely, in the same example, the priority might be to enhance the confidence in the semantic feature of the classifier in order to call the police and stop the loss due to burglary without the need for a better resolution. In this case, one can design a head by setting $\alpha_{22} = 0$.

The parameters $\bm{\beta}_{lk}$ are employed to model the significance of the accuracy of the semantic extractor in identifying different classes. Each element of the diagonal matrix $\bm{\beta}_{lk}$ is between zero (lowest priority) and one (highest priority). We will see in Section~\ref{sec:Exp_section}, that the selection of different values of the diagonal elements of $\bm{\beta}_{lk}$ will affect recall and precision in different ways. This is due to the fact that the cross entropy term in semantic loss of \eqref{eq:loss} only tries to maximize the sum of the product of the prior and the posterior probabilities for all classes, which does not correspond to the optimization of either of recall or precision. However, a suitable choice for $\bm{\beta}$ will help to effectively balance both recall and precision as different accuracy metrics. 

Two notes are in order here. First, the semantics may include the segmentation or class labels of the input signal. In practice, the class labels have a hierarchical structure, i.e., 
$$\forall c_{lk} \in C_{lk}, \exists c_{l-1,k'} \in C_{l-1,k'}: c_{lk} \subset c_{l-1,k'}$$ 
where $C_{lk}$ is the set of class labels for the head $k$ of layer $l$. Second, the length of the vector $\hat{y}_{lk}$, $\mathbbm{1}(y_{lk}(x))$, and the order of the square matrix $\bm{\beta}_{lk}$ is equal to the size of the set $C_{lk}$.

Total loss $\mathcal{L}$ is a weighted sum of losses calculated by all layers and all heads inside each layer, namely $\mathcal{L}_{lk}$, 
\begin{align} 
    \mathcal{L} = \sum_{l=1}^L \sum_{k=1}^{K_l} c_{lk} \mathcal{L}_{lk}
\end{align}

The sub-block size $n_l$ and weight $c_{lk}$ may be used to tune the performances for different resolutions or accuracy obtained by different encoder heads or sub-blocks. Moreover, the training process may benefit from adaptive adjustment of the weights $c_{lk}$ to enforce faster convergence for some encoder heads prior to optimization of the other heads. For example, this approach may be used to design the differential resolution obtained by an encoder head in layer $l$ conditional on a threshold performance by previous layers.  

We use the loss function \eqref{eq:loss} to train the encoder-decoder pair. For faster convergence and ensure low semantic loss from the beginning, the semantic extractors are initially pre-trained with raw data. We use alternative optimization to update the semantic extractors after training the encoder-decoder pair in each epoch. As a result, the semantic extractors will be trained to provide the best accuracy based on the reconstructed data instead of the raw data, hence, incorporating the effect of the designed AE and channel characteristics.

\section{Experimental Results} \label{sec:Exp_section}

\subsection{Dataset Specifications}
We use two datasets in our experiments: I) MNIST dataset that consists of 28 × 28 gray-scale images of handwritten digits categorized into 10 classes. The training (test) set contains 60,000 (10,000) images. II) The CIFAR-10 dataset that consists of 32 x 32 RGB images divided into 10 classes with each class representing a specific object category, such as airplanes, automobiles, birds, cats, deer, dogs, frogs, horses, ships, and trucks. The training (test) set contains 50,000 (10,000) images.

\begin{figure}[]
 \centerline{\includegraphics[width=0.7\linewidth]{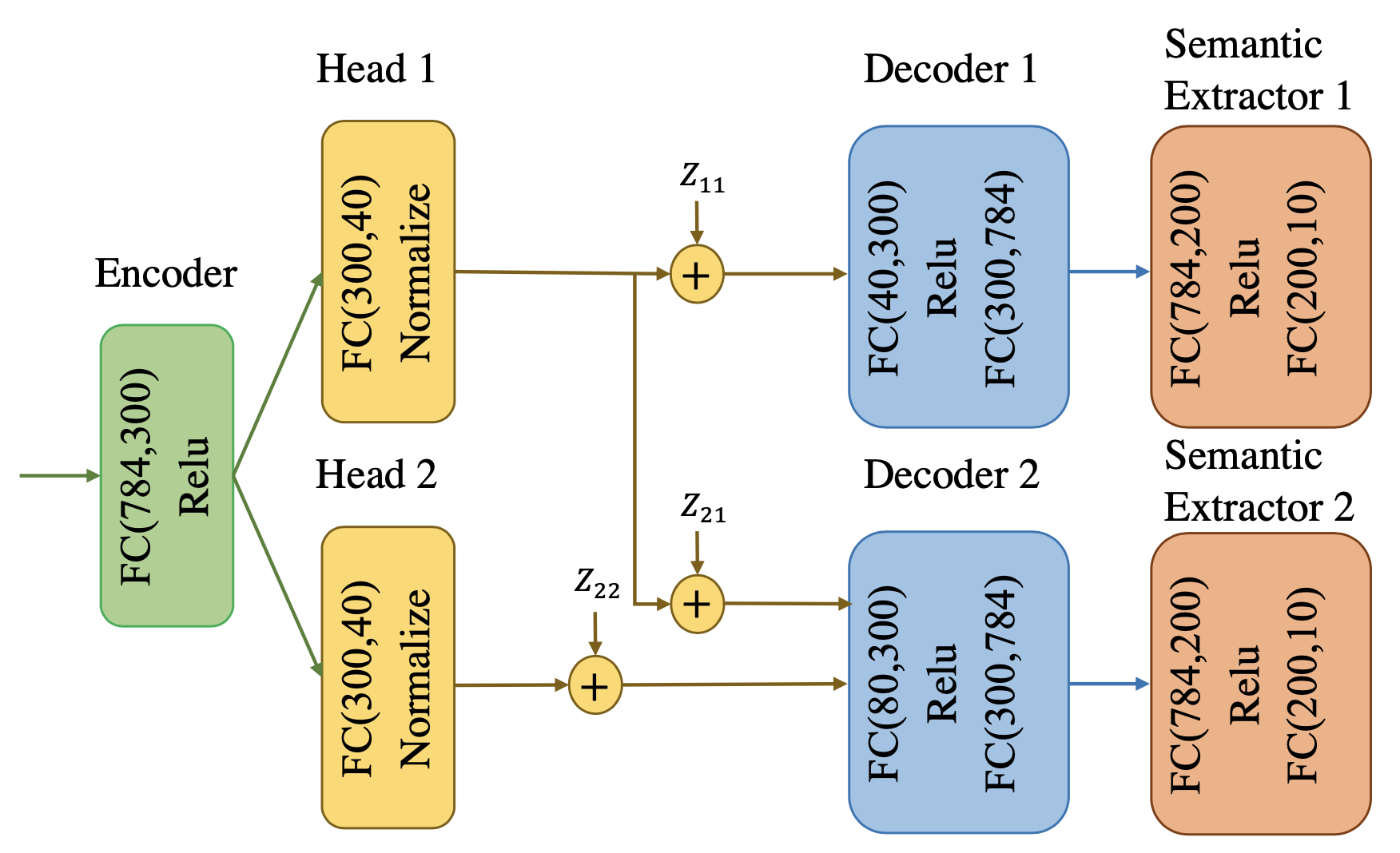}}
  \caption{Model structure for MNIST dataset.}
  \label{MNIST_model}
  \vspace*{-0.4cm}
\end{figure}

\begin{figure}[]
 \centerline{\includegraphics[width=0.7\linewidth]{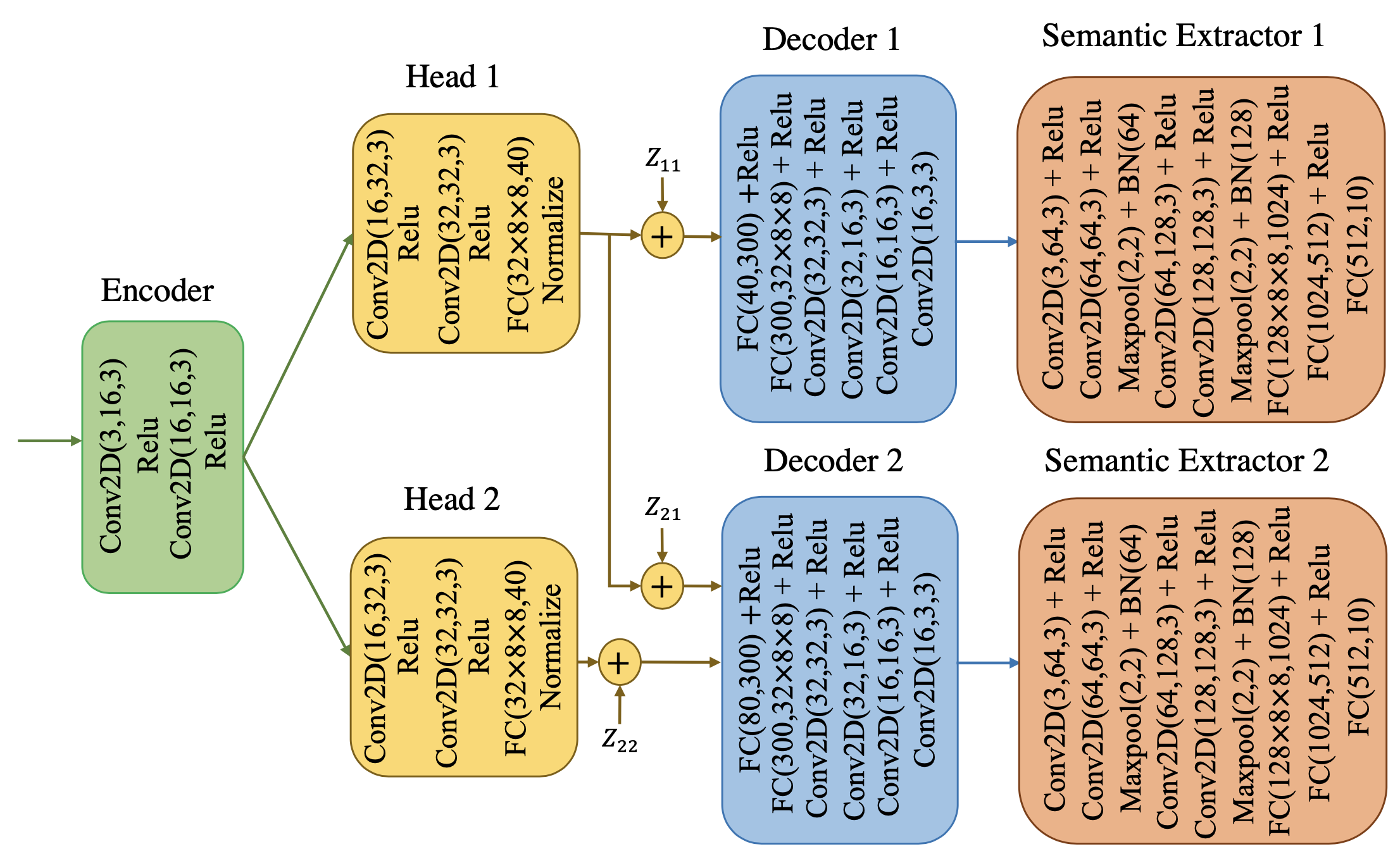}}
  \caption{Model structure for CIFAR10 dataset.}
  \label{CIFAR_model}
  \vspace*{-0.4cm}
\end{figure}

\subsection{Model and Hyperparameters}
We use Adam optimizer with learning rate $ \eta = 1 \times 10^{-3}$ to individually train our proposed SMRC parameters for both datasets. For MNIST dataset, the model for encoder, headers, and decoders can be seen in Fig.~\ref{MNIST_model}. For CIFAR-10 dataset, the models can be seen in Fig.~\ref{CIFAR_model}. We use equal weights for all $c_{lk}, \forall l, k$. Also, we set $\alpha_{lk} = 0.9, \forall l, k$. In all the simulations, we consider SMRC to a single user, i.e., we set $z_{ki} = z_{li}, \forall k,l \in [L]$.
\subsection{Reconstruction Performance (RP)}
We evaluate the reconstruction performance in terms of peak signal-to-noise ratio (PSNR), 
\begin{align}
    \text{PSNR (dB)} = 10\log_{10}\left(\frac{\max^2}{MSE}\right).
\end{align}

\begin{figure}[]
 	\begin{center}
 	\subfigure[]{%
 	\includegraphics[width=0.49\linewidth]{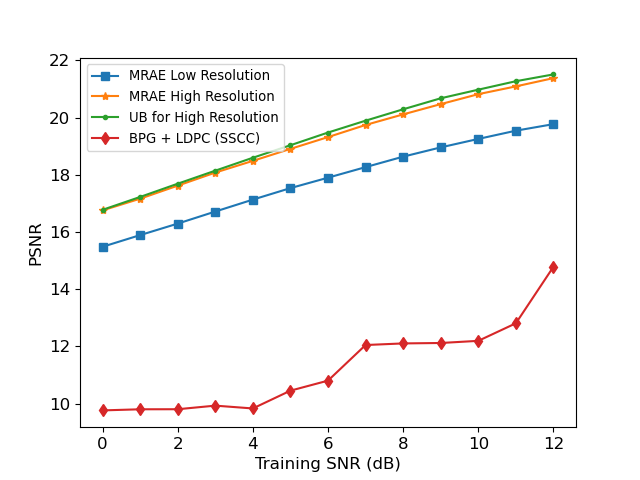}}
 	\subfigure[]{%
 	\includegraphics[width=0.49\linewidth]{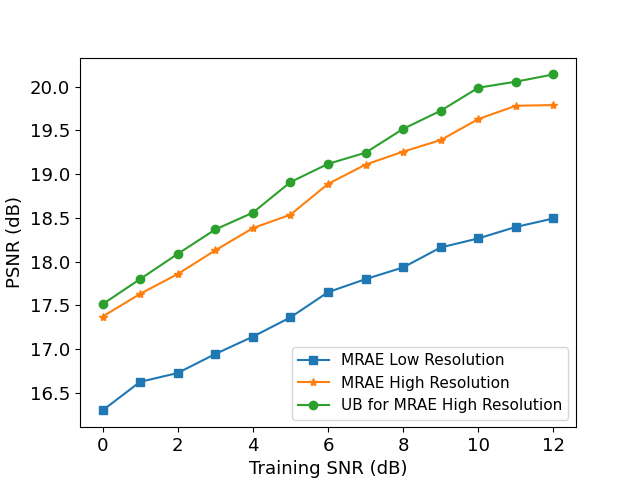}}\\ 
 	\end{center}
    \vspace*{-0.4cm}
 	\caption{PSNR results for (a) MNIST and (b) CIFAR10 dataset.}
 	\label{MNIST_PSNR}
    \vspace*{-0.4cm}
\end{figure}

We employ SMRC as in Fig.~\ref{MNIST_model} to encode an image $x$ from MNIST dataset and generate an encoded sub-block of size 40 symbols by each head. The first decoder receives the first sub-block plus noise and reconstructs the image as $\hat{x}_1$, while the second decoder utilizes both sub-blocks to generate $\hat{x}_2$. 
Fig.~\ref{MNIST_PSNR}(a) illustrates the PSNR $\hat{x}_1$ and $\hat{x}_2$ versus the SNR of the communication channel (denoted by the training SNR). It can be seen that in comparison to the first decoder, the second decoder improves PSNR by about 2 dB for SNR in the range of 0 to 12 dB. We also plot the RP of single head AE with the same complexity as our two head SMRC encoder that generates an output block of size 80 symbols directly. This will serve as an upper bound on the performance of the second decoder of SMRC with two heads. Fig.~\ref{MNIST_PSNR}(a) shows that the second decoder of SMRC has negligible loss in comparison to the upper bound that reveals the effectiveness of the successive encoding by SMRC.

In order to assess the performance of our framework in comparison to a SSCC, we employ the BPG (Better Portable Graphics) algorithm as the source coder and LDPC (Low-Density Parity-Check) codes for the channel coding along with MCS tables from 3GPP/LTE standard (Table 7.2.3 in \cite{3gpp_TS36_213}). Fig.\ref{MNIST_PSNR}(a) shows that using the same number of channel uses, i.e., 80 symbols, our proposed methods consistently outperform SSCC (comprising BPG as a source coding and LDPC as a channel coding scheme) by a significant margin.

We replicate the experiment with CIFAR10 dataset in Fig.~\ref{MNIST_PSNR}(b). The results again show that our framework improves output quality successively and approaches the upper bound obtained by training a single head AE that encodes the image with 80 symbols. 

\subsection{Semantic Accuracy (SA)}
To demonstrate the efficacy of our framework in preserving intended semantic information, we utilize two key accuracy metrics with respect to a classifier: recall and precision. 
In the multi-class scenario, we define the accuracy in terms of per-class recall and precision. We define \emph{recall for class $i$} as $\rho_i \triangleq P(\hat{y} = i|y(x) = i)$, that is the probability of correctly classifying the class of the output $\hat{x}$, i.e., $\hat{y}$, as $i$ given that the class label of the input signal $x$, i.e., $y(x)$ is $i$.  Also, we define \emph{precision} as  $\pi_i \triangleq P(y(x) = i|\hat{y} = i)$ that is the probability of the input label $y(x)$ being equal to $i$ given that the class label of the output, i.e., $\hat{y}$, is equal to $i$. However, we define \emph{average accuracy} as the average recall for a given input distribution function, i.e., $\sum_i \rho_i P(y(x) = i)$, that is equal to average precision.

\begin{figure}[]
 \centerline{\includegraphics[width=0.9\linewidth]{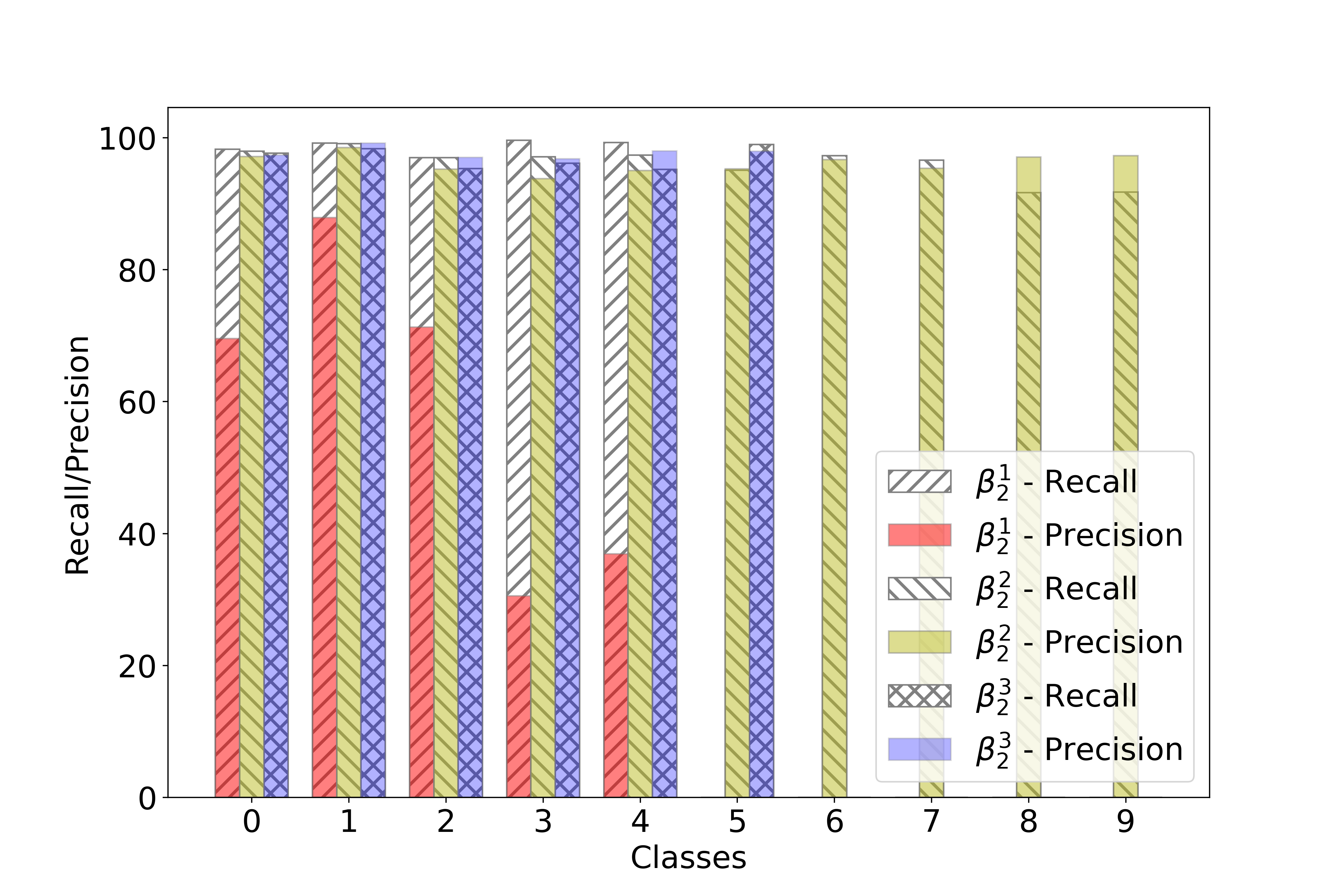}}
  \vspace*{-0.3cm}
  \caption{Effect of $\beta$ (weight vector) in precision/recall.}
  \label{prec_recall}
  \vspace*{-0.4cm}
\end{figure}

We evaluate SA for the same two-layer SMRC used for RP evaluation. Using MNIST dataset, we train the first layer encoder to classify all 10 classes, while the second layer encoder is designed to improve the recall or precision of the first 5 classes by utilizing a weighting vector $\beta$ in the cross-entropy loss function (\ref{eq:loss}). Fig.~\ref{prec_recall} illustrates the recall and precision outcomes obtained for some choices of the weighting vector.
In the first case, denoted as $\beta_2^1$, we use weight 1 for each of the first 5 classes and 0 for the rest. In the second weighting choice, $\beta_2^2$, we use equal weight 1 for the first 5 classes and equal weight $1/2$ for the second 5 classes. Finally, in the third weighting scheme $\beta_2^3$, we assign weight 1 for the first 5 classes and pool the second 5 classes into one ``dummy'' class with the weight 1.

These different weighting schemes allow us to assess the impact of varying the importance assigned to different classes in the second layer. Fig.~\ref{prec_recall} depicts the recall (precision) as a measure of accuracy of second layer output for each of the used weighting vector in cross-entropy loss. 
Using $\beta_2^1$ results in the best improvement in recall of successive layer, but it comes at the cost of degrading the precision since the encoder mistakenly learns to map the rest of non-important classes (last 5) into the important classes (first 5). This can be mitigated by choosing $\beta_2^2$ which restores the precision to an acceptable level while still improves the recall performance of the second layer. However, the recall performance with $\beta_2^2$ is not as good as that of $\beta_2^1$. Finally, by mapping all unimportant classes into a single class, using $\beta_2^3$ provides the best precision performance for the chosen classes at the expense of slightly lower recall than that of $\beta_2^1$ or $\beta_2^2$. 

\begin{figure}[]
 	\begin{center}
 	\subfigure[]{%
 	\includegraphics[width=0.49\linewidth]{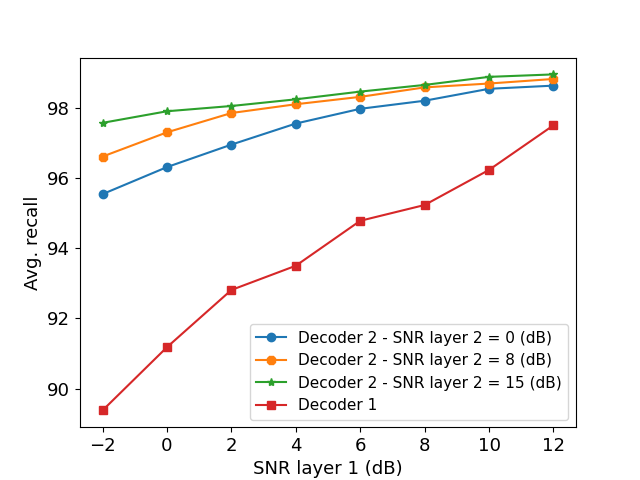}}
 	\subfigure[]{%
 	\includegraphics[width=0.49\linewidth]{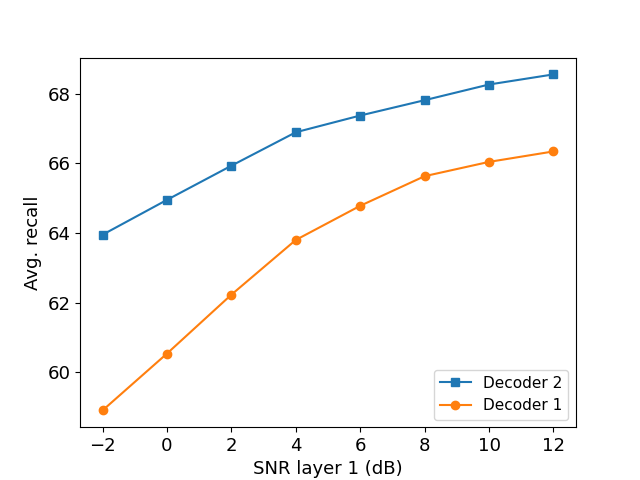}}
 	\end{center}
    \vspace*{-0.4cm}
 	\caption{Accuracy (avg. recall) of different layer decoders in different SNR for (a) MNIST (b) CIFAR10 dataset.}
 	\label{Recall_MNIST_CIFAR10}
\end{figure}

The average recall performance of the reconstructed outputs as a function of the first layer channel gain is illustrated in Fig.~\ref{Recall_MNIST_CIFAR10}(a) for MNIST dataset. We further evaluate the performance of the second layer decoder output across three distinct second layer channel gains: 0 dB, 8 dB, and 15 dB. The results consistently demonstrate that the second decoder output exhibits superior recall performance. We run the same evaluation for CIFAR10 dataset. Fig.~\ref{Recall_MNIST_CIFAR10}(b) again shows that the successive layer improves the classification performance for CIFAR10 dataset.

\bibliographystyle{unsrt}
\bibliography{reference}

\begin{thebibliography}{10}

\bibitem{shannon1948mathematical}
C.~E. Shannon.
\newblock A mathematical theory of communication.
\newblock {\em The Bell system technical journal}, 27(3):379--423, Oct 1948.

\bibitem{gunduz2009source}
D.~Gunduz, E.~Erkip, A.~Goldsmith, and H.~V. Poor.
\newblock Source and channel coding for correlated sources over multiuser
  channels.
\newblock {\em IEEE Transactions on Information Theory}, 55(9):3927--3944, Sep
  2009.

\bibitem{kurka2020deepjscc}
D.~B. Kurka and D.~G{\"u}nd{\"u}z.
\newblock Deepjscc-f: Deep joint source-channel coding of images with feedback.
\newblock {\em IEEE Journal on Selected Areas in Information Theory},
  1(1):178--193, May 2020.

\bibitem{yang2021deep}
M.~Yang, C.~Bian, and H.~Kim.
\newblock Deep joint source channel coding for wirelessimage transmission with
  ofdm.
\newblock Available online at arXiv:2101.03909.

\bibitem{yang2023witt}
K.~Yang, S.~Wang, J.~Dai, K.~Tan, K.~Niu, and P.~Zhang.
\newblock Witt: A wireless image transmission transformer for semantic
  communications.
\newblock In {\em IEEE ICASSP}, pages 1--5, June 2023.

\bibitem{baldi2012autoencoders}
P.~Baldi.
\newblock Autoencoders, unsupervised learning, and deep architectures.
\newblock In {\em ICML}, pages 37--49, June 2012.

\bibitem{aoudia2019model}
F.~A. Aoudia and J.~Hoydis.
\newblock Model-free training of end-to-end communication systems.
\newblock {\em IEEE Journal on Selected Areas in Communications},
  37(11):2503--2516, Nov 2019.

\bibitem{aoudia2021end}
F.~A. Aoudia and J.~Hoydis.
\newblock End-to-end learning for ofdm: From neural receivers to pilotless
  communication.
\newblock {\em IEEE Transactions on Wireless Communications}, 21(2):1049--1063,
  Feb 2021.

\bibitem{liu2023multiresolution}
Y.~Liu, C.~Ponce, S.~L. Brunton, and J.~N. Kutz.
\newblock Multiresolution convolutional autoencoders.
\newblock {\em Journal of Computational Physics}, 474:111801, Feb 2023.

\bibitem{kurka2019successive}
D.~B. Kurka and D.~G{\"u}nd{\"u}z.
\newblock Successive refinement of images with deep joint source-channel
  coding.
\newblock In {\em IEEE SPAWC}, pages 1--5, July 2019.

\bibitem{sagduyu2022semantic}
Y.~E. Sagduyu, T.~Erpek, S.~Ulukus, and A.~Yener.
\newblock Is semantic communications secure? {A} tale of multi-domain
  adversarial attacks.
\newblock Available online at arXiv:2212.10438.

\bibitem{sagduyu2023vulnerabilities}
Y.~E. Sagduyu, T.~Erpek, S.~Ulukus, and A.~Yener.
\newblock Vulnerabilities of deep learning-driven semantic communications to
  backdoor (trojan) attacks.
\newblock In {\em IEEE CISS}, pages 1--6, March 2023.

\bibitem{Zhang2017ASO}
Y.~Zhang and Q.~Yang.
\newblock A survey on multi-task learning.
\newblock Available online at arXiv:1707.08114.

\bibitem{3gpp_TS36_213}
3GPP~TS36.213 V12.3.0.
\newblock Evolved universal terrestrial radio access (e-utra); physical layer
  procedures.

\end{thebibliography}
\end{document}